\documentclass{article}

\usepackage{PRIMEarxiv}

\usepackage[utf8]{inputenc} 
\usepackage[T1]{fontenc}    
\usepackage{hyperref}       
\usepackage{url}            
\usepackage{booktabs}       
\usepackage{amsfonts}       
\usepackage{nicefrac}       
\usepackage{microtype}      
\usepackage{lipsum}
\usepackage{fancyhdr}       
\usepackage{graphicx}       
\graphicspath{{media/}}     

\title{Long Story Generation via Knowledge Graph and Literary Theory}

\author{
  Ge Shi, Kaiyu Huang, Guochen Feng \\
  Beijing Jiaotong University, Beijing, China \\
  Beijing, China\\
  \texttt{\{23121732, kyhuang, gcfeng\}@bjtu.edu.cn} 
} 

\begin{document}
\maketitle

\begin{abstract}
The generation of a long story consisting of several thousand words is a sub-task in the field of long text generation~(LTG). Previous research has addressed this challenge through outline-based generation, which employs a multi-stage method for generating outlines into stories. However, this approach suffers from two common issues: almost inevitable theme drift caused by the loss of memory of previous outlines, and tedious plots with incoherent logic that are less appealing to human readers. \\

In this paper, we propose the multi-agent Story Generator structure to improve the multi-stage method, using large language models~(LLMs) as the core components of agents. To avoid theme drift, we introduce a memory storage model comprising two components: a long-term memory storage that identifies the most important memories, thereby preventing theme drift; and a short-term memory storage that retains the latest outlines from each generation round. To incorporate engaging elements into the story, we design a story theme obstacle framework based on literary narratology theory that introduces uncertain factors and evaluation criteria to generate outline. This framework calculates the similarity of the former storyline and enhances the appeal of the story by building a knowledge graph and integrating new node content. Additionally, we establish a multi-agent interaction stage to simulate writer-reader interaction through dialogue and revise the story text according to feedback, to ensure it remains consistent and logical. Evaluations against previous methods demonstrate that our approach can generate higher-quality long stories.
\end{abstract}


\section{Introduction}
\begin{figure}
    \centering
    \includegraphics[width=0.5\linewidth]{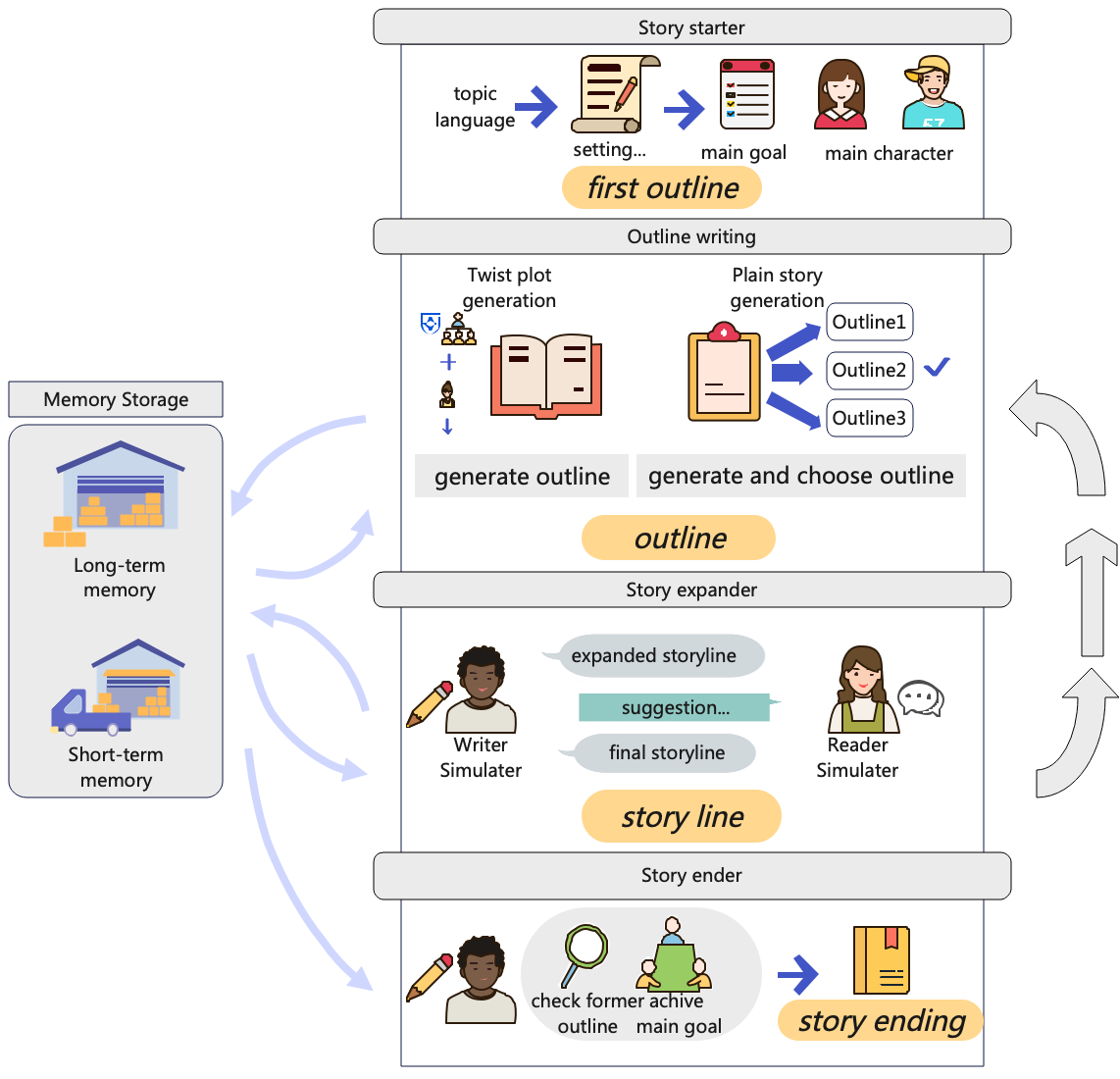}
    \caption{Overview of the Story Generator structure.}
    \label{fig1}
\end{figure}
In the field of long text generation~(LTG), long story generation has emerged as a general yet challenging sub-task. Automating the generation of long stories can have far-reaching applications, such as in the entertainment industry for creating engaging scripts~\cite{bae-kim-2024-collective}, in the education field for developing interactive learning materials~\cite{jiang-etal-2024-leveraging}, and in the realm of virtual reality for constructing immersive narrative experiences~\cite{10877467}. However, the task of generating long stories is far from trivial. For single-agent generators driven by LLMs that have not been fine-tuned, their automatic story generation performs disappointingly when generating without structured and efficient instructions~\cite{tian-etal-2024-large-language}. Unlike short stories, long stories' generation should maintain coherence and consistency over several thousand words. Thus, the single-agent generation always shows limitations in its results. It must ensure that the plot unfolds logically, characters’ actions are motivated, and the overall theme is well developed throughout the narrative~\cite{yang2024makesgoodstorymeasure}. The model needs to comprehensively master language knowledge, world knowledge, and commonsense knowledge to create a story that is both meaningful and engaging to human readers~\cite{gao2025llm}.

To instruct story generation, previous research has primarily adopted outline-based generation methods to tackle the long story generation challenge~\cite{yang-etal-2023-doc}. These methods typically follow a multi-stage process from outline to story. Despite some achievements, they are plagued by several common issues.

One of the most prominent problems is the almost inevitable theme drift~\cite{xie-riedl-2024-creating}. As the generation process progresses, the model often loses memory of the previously set outline, which leads to situations where the story suddenly veers off-track, with plot elements that are inconsistent with the initial premise or overall theme. For example, in a story initially set in a medieval fantasy world, the model might introduce modern technological elements without proper justification~\cite{yang2024makesgoodstorymeasure,gao2025llm}.

Another significant drawback is the production of dull and unappealing plots~\cite{tian-etal-2024-large-language}. The generated stories often lack the twists, turns, and emotional depth that are characteristic of good human-written stories~\cite{doshi2024generative,yang-etal-2023-doc}. The narratives may be overly simplistic, with a predictable sequence of events that fail to capture the reader's attention or evoke any strong emotions~\cite{bae-kim-2024-collective}.

To address the above limitations, this paper proposes a “Story Generator” structure. Our approach represents a significant departure from previous methods and also offers several novel contributions. We follow literary theory to create appealing plots for the story by introducing new nodes into the story's KG structure, and using multi-agent dialogues to improve the expanded story's statement. To make better choices in the twists and regular narration of the story, we use the similarity calculation method of the former outlines to automatically determine which outline generation strategy to use.

\begin{itemize}
    \item We introduce a memory storage with two components. The long-term memory storage identifies the most important memories which effectively anchors the story to the original theme. The short-term memory storage preserves the last two outlines in each generation round, providing immediate context for the ongoing generation.
    \item We design a framework for generating story twists based on narratology theory to enhance the attractiveness of the story. By constructing a knowledge graph and integrating new node content, the story introduces new conflicts and factors to make it more interesting to read.
    \item We establish a multi-agent system to simulate the interaction between writer and reader through dialogue. Agents generate, review, and edit the story texts, enabling revisions to ensure that stories are easy to understand and follow the basic logic.
\end{itemize}
Overall, our Story Generator structure has the potential to generate long stories of higher quality, bridging the gap between LLM-generated and human-written stories.

\section{Related Work}
\subsection{Long Text Generation}
Before the era of Large Language Models (LLMs), long text generation, including long story generation, was a daunting challenge. Researchers typically relied on Long Short-Term Memory (LSTM) models~\cite{hochreiter1997long} to predict the probability of a token appearing. However, this approach was inefficient and yielded suboptimal results, as LSTMs struggled to capture long range dependencies in texts spanning thousands of words, often leading to fragmented narratives and inconsistent logic~\cite{sutskever2014sequence}.

With LLMs leading the way, NLP-assisted LTG has gradually become the mainstream method. Ideally, using LLMs specifically designed for LTG would be the optimal solution. For example, models like Longformer ~\cite{Beltagy2020LongformerTL}
 and FlashAttention~\cite{dao2022flashattention} incorporate innovative attention mechanisms to handle extended contexts efficiently. However, such methods typically demand large memory capacity and high-performance GPUs. If using fine-tuned LLMs is not an option, the design of instructions for LTG becomes crucial, the second-best solution is to depend on external computing, for example, using API models such as GPTs. Recent studies have emphasized the importance of prompt engineering in guiding LLMs to produce structured long texts.
\subsection{Multi-agent structure}

In the current research on LTG, where multi-agent systems are a primary approach, a major challenge lies in structuring the generated text. Frameworks like Agents' Room~\cite{choi-etal-2025-agents} highlight that without clear structural guidelines, LTG tends to become disorganized and illogical. This aligns with findings from chain of agents~\cite{zhang2024chain} emphasize that explicit structural instructions are vital for maintaining coherence in multi-agent-generated long-form texts.

Early solution approaches relied on multi-stage frameworks that decomposed the story generation process into sequential steps, such as outline creation followed by story expansion. For instance, the DOC framework~\cite{yang-etal-2023-doc} improved plot coherence by enforcing detailed hierarchical outlines, shifting creative responsibility from the drafting stage to the planning phase. However, these methods often suffered from theme drift due to limited contextual retention, as models struggled to maintain consistency across thousands of words.
\begin{figure*}[t]
\centering
\includegraphics[width=0.8\linewidth]{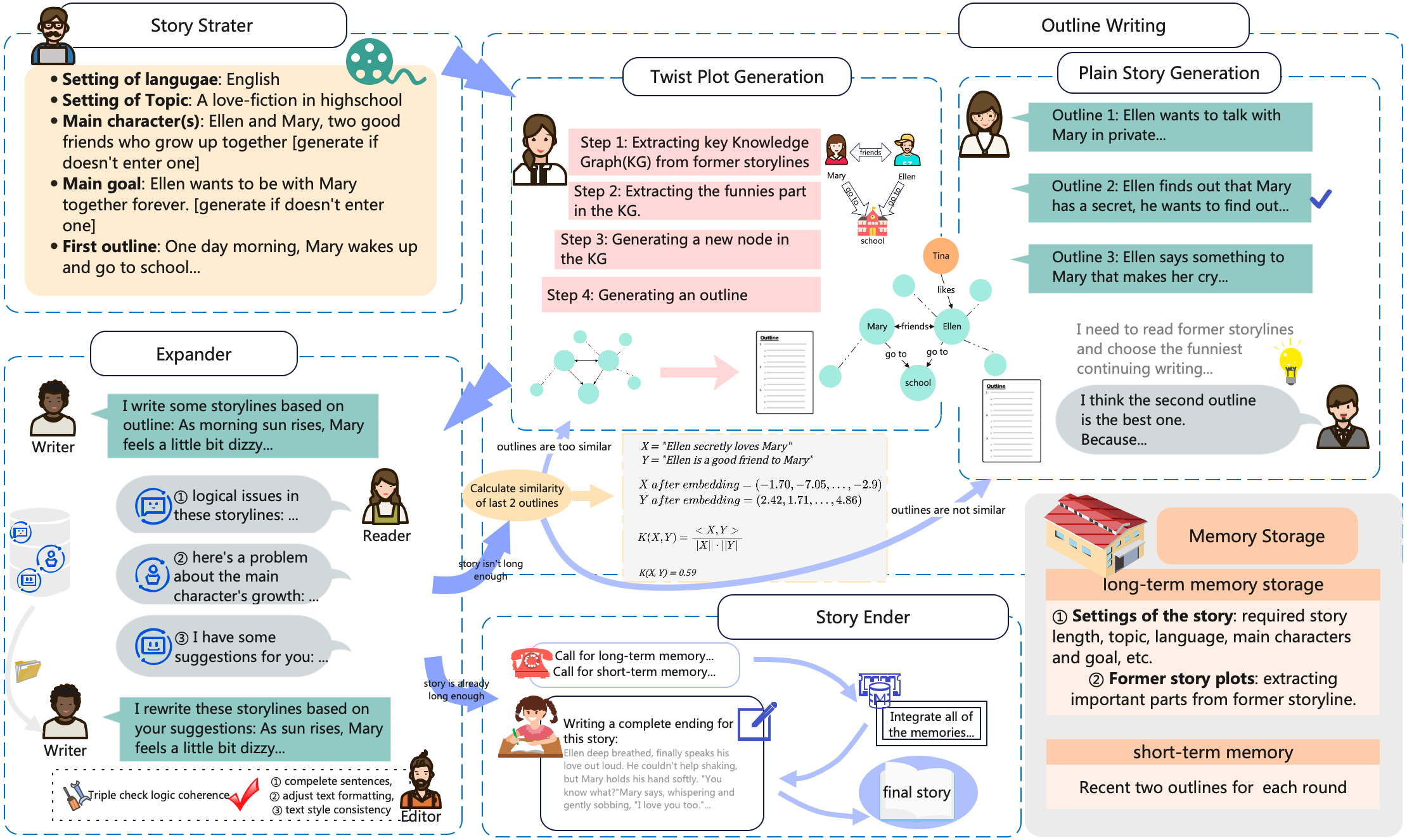} 
\caption{The structure of Story Generator. Begins in the Story Starter stage, then Outline Writing. Expender expands outlines and triple-checks storylines. Story Ender gives a good ending to the story. Memory Storage keeps following the whole generation process and gives back responses when Generators call it.}
\label{fig2}
\end{figure*}

\subsection{Memory Mechanisms in NLP}
Another challenge of multi-agent driven long story generation is the Lost in the Middle phenomenon~\cite{liu-etal-2024-lost} discovered through experiments that when LLMs receive long inputs, they tend to focus on the information at the beginning and end of the input sequence while neglecting the content in the middle. This phenomenon can cause the model to miss crucial information, resulting in a lack of coherence and logicality in the LTG. For example, in a long story generation task, the model might forget the character traits or plot points introduced in the middle part of the story, leading to inconsistent character behaviors or plot development.

To address this, memory-augmented architectures emerged in multi-agent driven LTG. The DOME~\cite{wang-etal-2025-generating} introduced a dynamic hierarchical outlining mechanism combined with a temporal knowledge graph to store and retrieve critical narrative elements, reducing contextual conflicts and enhancing long-range coherence. Similarly, MemLong~\cite{liu2024memlongmemoryaugmentedretrievallong} leveraged external memory retrieval to extend context lengths up to 80k tokens, mitigating the quadratic complexity of traditional attention mechanisms. These advancements highlighted the importance of structured memory systems in preserving thematic consistency.

\subsection{Literary Theories in Story Generation}
Another critical challenge lies in narrative engagement. A study indicated that one of the most difficult and complex tasks in evaluating a good storyteller is assessing the interesting-ness of their narratives~\cite{yang2024makesgoodstorymeasure}. Uninstructed methods often produce predictable plots lacking emotional depth.
 
As E.M. Forster~\cite{forster1927aspects} mentioned in Aspects of the Novel, enriching story generation by distinguishing between story (a chronological sequence of events) and plot (a sequence driven by causal relationships) is a brilliant way to interest readers. Forster’s concepts of flat characters (one-dimensional, predictable figures) and round characters (complex, evolving personalities) also guide nuanced character development. This distinction helps avoid simplistic characterizations: while flat characters may serve specific functional roles (e.g., a loyal sidekick), round characters that have conflicting motivations and growth arcs add depth and emotional resonance, making stories more engaging. For example, a round protagonist grappling with moral dilemmas can evoke stronger reader empathy than a static, one-note hero.

Structural closure is also a fundamental theoretical framework in narratology, whose core lies in the causal closure of narratives, that is, accidents must be incorporated into an intelligible logical chain. This theory has been widely confirmed by the academic community of narrative studies, represents include: Morphology of the Folktale \cite{propp1968morphology}, The Two Principles of Narrative~\cite{Todorov1971The2P}, and The Political Unconscious: Narrative as a Socially Symbolic Act~\cite{jameson1981political}.
Otherwise, the narrative will lose its meaning. For example, in detective novels, the occurrence of a case (an accident) necessarily leads to the revelation of the truth (resolution), in Bildungsromans, the protagonist's cognitive conflicts (accidents) ultimately achieve reconciliation through self-awakening. This structure is not only applicable to fictional narratives but also widely exists in various forms, such as folktales and films. Whether  which method of narrative, structural closure has always been a crucial hallmark of legitimacy.

\section{Story Generator Architecture}

\subsection{Overview of Story Generator}

In this section, we'll introduce the process of  the Story Generator multi-agent structure unfolds in four key stages, as we show in Figure~\ref{fig2}.

In the Story Starter phase, based on the given topic and language, generate the settings, including the story's main character or characters, and their main goal. The settings are defined to form the initial first outline and laying the background for the story.

In the Outline Writing stage, LLM-driven agents attempt to continue writing an outline. Two parallel processes occur. One of the twists spot generation and plain story generation works: the former injects twists and produces an outline, the latter keeps writing outline without forcing twists. These are then filtered via a selection mechanism to pick the one. Thus, the outline-generating structure could continuously refine the story framework.

The Story Expander stage features interaction between the writer simulator and reader simulator. The writer simulator first drafts an expanded storyline, and the reader simulator provides feedback. Their interaction yields storylines, making the plot richer and more logically consistent. After expanding, Story Generator calculates the length of this whole story. If the story isn't long enough,  it goes back to the Outline Writing stage. If the story is long enough, it goes to the final step below.

Finally, in the Story Ender stage, a check of the initial outline and main goals ensures the ending aligns with the overall theme. Backed by a memory storage model~(preventing theme drift), a story theme obstacle framework~(boosting appeal with KG-driven twists), and multi-agent interaction~(revising via feedback), this process overcomes issues like theme drift in traditional methods, enabling the creation of high-quality long stories. In addition to the mentioned structure above, throughout the entire process, a long-term memory storage continuously follows and records important information, a short-term memory storage retains the generated content of the last two rounds.


\begin{figure*}[t]
\centering
\includegraphics[width=0.8\linewidth]{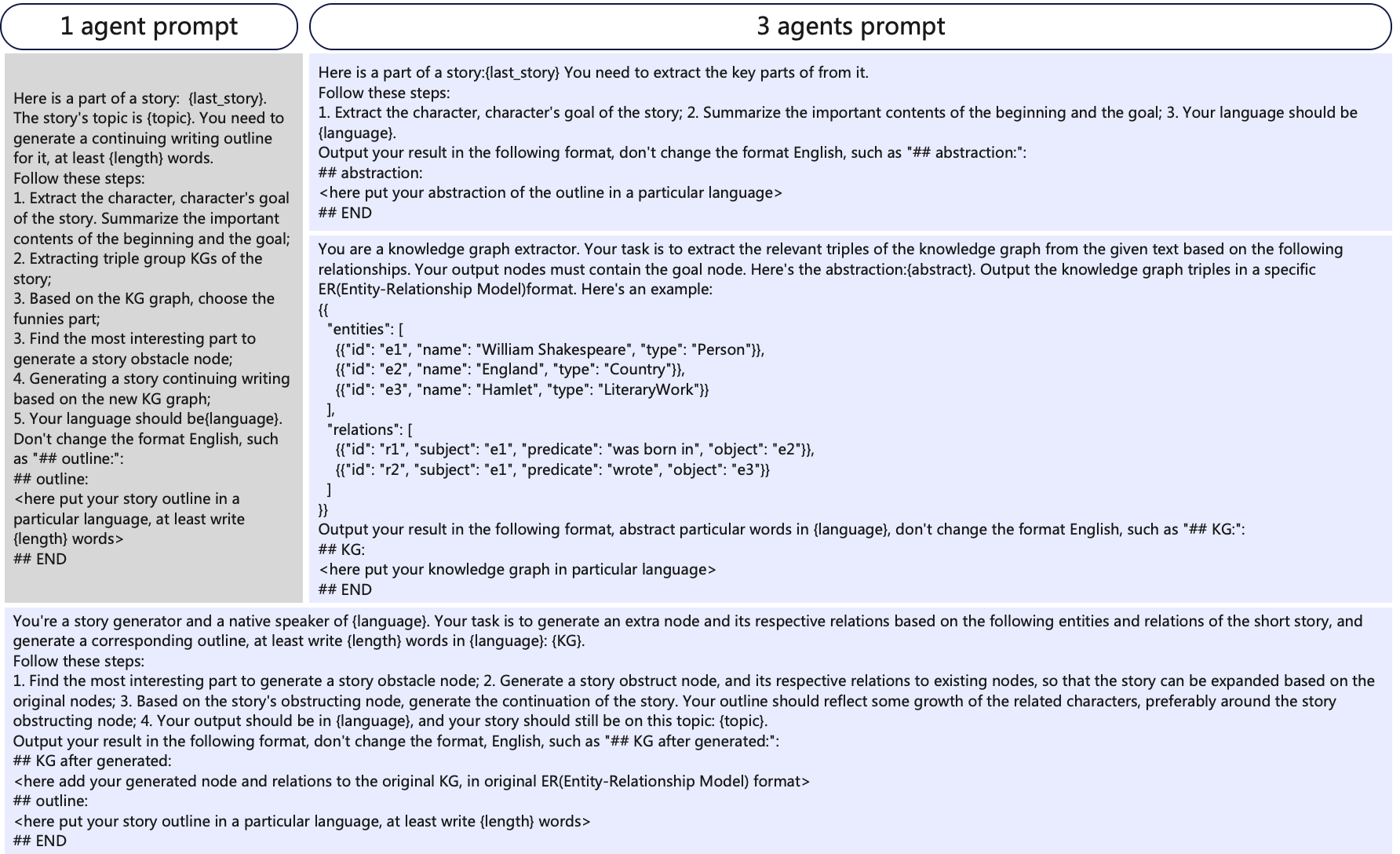} 
\caption{Prompts of two different KG-driven outline generation methods.}
\label{fig3}
\end{figure*}

\subsection{Memory Storage Strategy}
To minimize theme drift to the extent, we introduce long-term memory and short-term memory to collect important information generated during the outline and storyline generation process, and provide an interface for easy reading. The setup of long-term memory uses an LLM as the core module for text summarization. Long-term memory, firstly, records all the information generated by Story Starter, including the story's topic, main characters, their settings, the main goal, and a first outline. The output from each round of outline and text generation is fed into the LLM, and the LLM is instructed to extract important information and write it into the memory repository. As for short-term memory storage, it records the two most recently generated outlines. In the first round of generation, the memory holds the first-segment outline. At each stage of outline generation and content expansion, the content in the long-term memory storage repository is retrieved to ensure that the generation at each stage does not deviate from the theme, in the mean time, the content in the short-term memory storage repository is retrieved to ensure that the text style and details of the generation match those of the preceding text. 

\subsection{Literary Theories Applying Strategy}
According to the reference introduced account of story and plot in Aspects of the Novel~\cite{forster1927aspects}, plot is an essential element of a story. Universal narratology holds that twists in a story are achieved through plot, and an interesting plot enhances the coherence and readability of the story. This is to some extent supported by the findings on similarity calculation of chunked texts in the $EX^3$ method~\cite{lei-etal-2024-ex3}. However, the vocabulary size generated by LLMs in a single instance is limited, even for large-parameter LLMs. We don't want new twists to emerge in the story every several segments of text.

Therefore, we introduce a outline similarity strategy, adopting the cosine similarity used in the $EX^3$ experiment to calculate the similarity between the latest two outlines. If the similarity between the latest two outlines is above the threshold, it indicates that a twist should be introduced in the recent story, and then a twist plot will be generated. Otherwise, a plain story outline, written by the plain story generator, will be employed.

We also hope that the generation of the twist plots follows the story's logic and the characters' relations, based on the structural closure theory, which emphasizes that the unfolding of a plot backed by its story should be related to the growth of the characters and the characters' recent states. To follow this rule, we use KG to obtain the current entities~(as nodes) and their relations~(as edges), then generate an external node to facilitate the generation of twist plots. In the story ender, we call for memories to ensure a closed loop of the story narration.

\subsection{KG-driven Twist Plots Strategy}
To add more interesting new content to the story while adhering to the basic logic of the story, we extract the KG related to the current main character's goal from the current story (the current goal is forced to be a node to prevent theme drift). Constructing a KG with the main character's current goal as the core node can effectively anchor the main line of the story. Then, a new obstacle node related to the main goal is generated. We specify the relationship between the newly generated node and the existing nodes, emphasizing that the content of the new node emerges as an obstacle. Finally, a new outline is written based on the newly generated KG.

\section{Experiments}
\subsection{Experimetal Setting}
The multi-agent driven LTG structure is complex and has many components. We want to explore its best performance as much as we can. We use GPT-3.5-turbo and Claude-3-7-sonnet to be its core LLMs, and test some important components to determine how to structure this generator.

We use English texts' generation as the tasks, choosing Auto-J~\cite{li2024generative} as the testing human simulator. Auto-J is a generative judge that has 6B and 13B versions designed to effectively evaluate the alignment of different LLMs with human preferences. In each task, we test with a set of genres of stories, including science fiction, romance, fantasy.

\subsection{Implementation Details}

In this paper, we adopt pair-wise comparison between baselines and the Story Generator, with human evaluation as the metric, following Yang et al.~\cite{yang-etal-2022-re3}.

\textbf{Task}: We evaluate all generated stories across six topics: science fiction, romance, fantasy, horror, suspense, and mystery. All outputs are instructed to have a length of at least 10,000~(10k) words.

\textbf{Metrics}: Human evaluation is a standard method for assessing story quality. While we have employed the Auto-J LLM-based evaluation for output texts already, real human judgment is incorporated to ensure comprehensive assessment, according to these five dimensions below:
\begin{itemize}
    \item \textbf{Interesting-ness}: Are the stories appealing and interesting enough for human readers?
    \item \textbf{Commonsense}: Are the stories logical and rational?
    \item \textbf{No theme drift}: For each one-story, do the beginning and ending share a same theme? Do scenes of the story share a same theme?
    \item \textbf{Relevant}: Do the stories consistent with the premise?
    \item \textbf{Readability}: Are the stories readable? 
\end{itemize}

\textbf{Baselines}: Although the Story Generator is designed for long-form generation, it is compared against several well-performing baselines and ablated variants. The selected generators are listed as follows:
\begin{itemize}
    \item \textbf{DOC}~\cite{yang-etal-2023-doc}: DOC is a frequently cited work in the field of story generation. We replaced the LLM core of DOC with GPT-3.5-turbo. Due to the inaccessible of GPT's weights, we bypassed the detailed controller. 
    \item \textbf{RECURRENTGPT} ~\cite{zhou2023recurrentgptinteractivegenerationarbitrarily}: RECURRENTGPT is a method inspired by LSTM and uses the recurrent blocks as the basic structure. It can output stories in the long-term, so we set it to directly generate 10k words at least.
    \item \textbf{$EX^3$}~\cite{lei-etal-2024-ex3}: $EX^3$ is a fine-tuning driven approach. To enhance the storytelling ability of the LLM itself, it divides the text into multiple segments and conducts three stages of work: Extracting, Excelsior, and Expanding. We have the model generate long texts, specifying that they should be no less than 10k words.
\end{itemize}

\begin{table*}[ht]
\centering

\label{tablename} 
    \begin{tabular}{lccccc}
    \hline
    &Interesting-ness& commonsense &No theme drift& Relevant&Readable\\
    \hline
         DOC&22.5&36.7&61.7&23.3& 11.7 \\
         Story Generator&\textbf{83.4}&\textbf{70.8} &\textbf{89.1}&\textbf{95.8}&\textbf{93.3}\\
         \hline
         RECURRENTGPT&28.3&32.5&0&27.5& 0 \\
         Story Generator&\textbf{69.2}&\textbf{80} &\textbf{92.5}&\textbf{63.3}&\textbf{88.3}\\
         \hline
         EX3&42.5&65.8&65.8&51.7& 60.8 \\
         Story Generator&\textbf{74.2}&67.5 &\textbf{73.3}&\textbf{85.8}&60\\
         \hline
    \end{tabular}
    \caption{Pair-wise comparison with baselines for 20 stories' generation~${(\%)}$. Since there are many types of stories, listing details is unnecessary so we only display the mean values of each topic. The horizontal axis represents different methods being compared, the vertical axis represents the metrics being compared. Table values indicate the percentage of annotators who believe the method’s novels align with the metrics. Results in different comparisons are not comparable with each other. Bold indicates significance with ${p<0.05}$.} 
    \label{table1}
\end{table*}

\subsection{Main Results}

As shown in Table~\ref{table1}. Pair-wise comparisons show Story Generator outperforms three baselines in most metrics. It scores higher than DOC in all five metrics, surpasses RECURRENTGPT notably in theme consistency and readability, and beats $EX^3$ in interesting-ness, theme drift, and relevance. It generates better long stories, validating its architecture.

\paragraph{Human evaluation.} For each topic, we generated 10 stories using each method. We recruited 20 people (native English speakers) from the prolific website and sent each pair of generated stories to them, asking them to evaluate whether they met the specified criteria. The questionnaire can be found in Appendix. Annotators are shown a story's settings and two corresponding stories. One by Story Generator, and another one by a different method.
They are asked to assess, under different metrics, which one is better, or if both are equally good or bad. Finally, the percentage of annotators who consider the stories generated by different methods to meet the corresponding metrics under the same experimental conditions is calculated. Each novel pairwise is evaluated by two annotators. On average, each person needs to evaluate 15 stories' pair combinations. In the actual process, the distribution of the tasks was not even.

\begin{figure}[h]
\centering
    \includegraphics[width=0.3\linewidth]{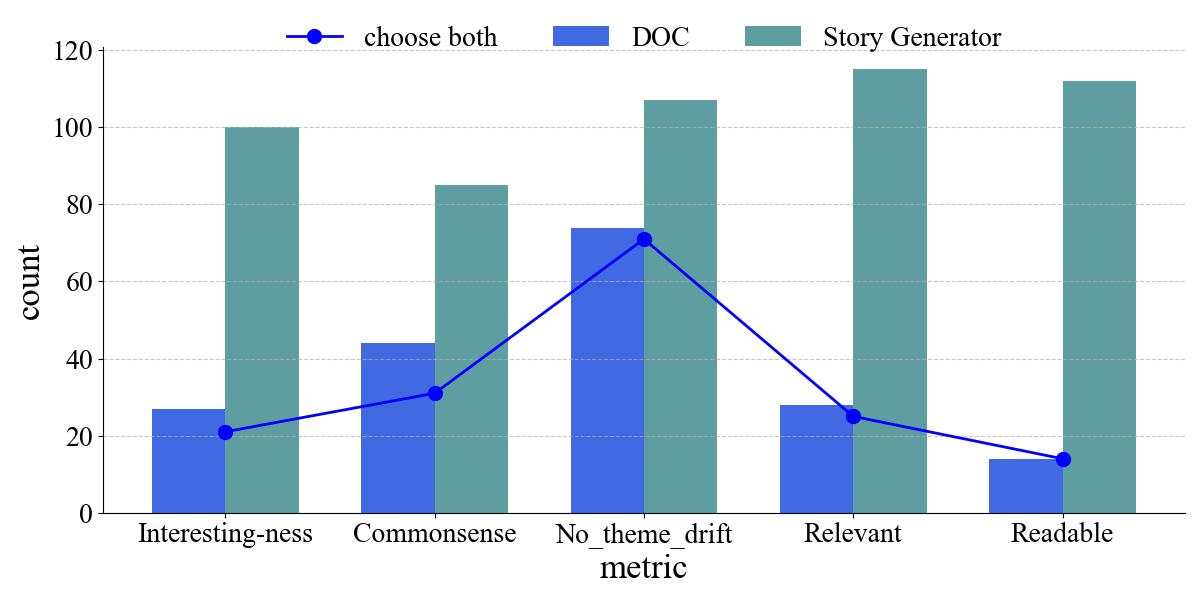}
    \caption{Comparing DOC and Story Generator. Counting identifies certain dimensions of the story.}

\centering
    \includegraphics[width=0.3\linewidth]{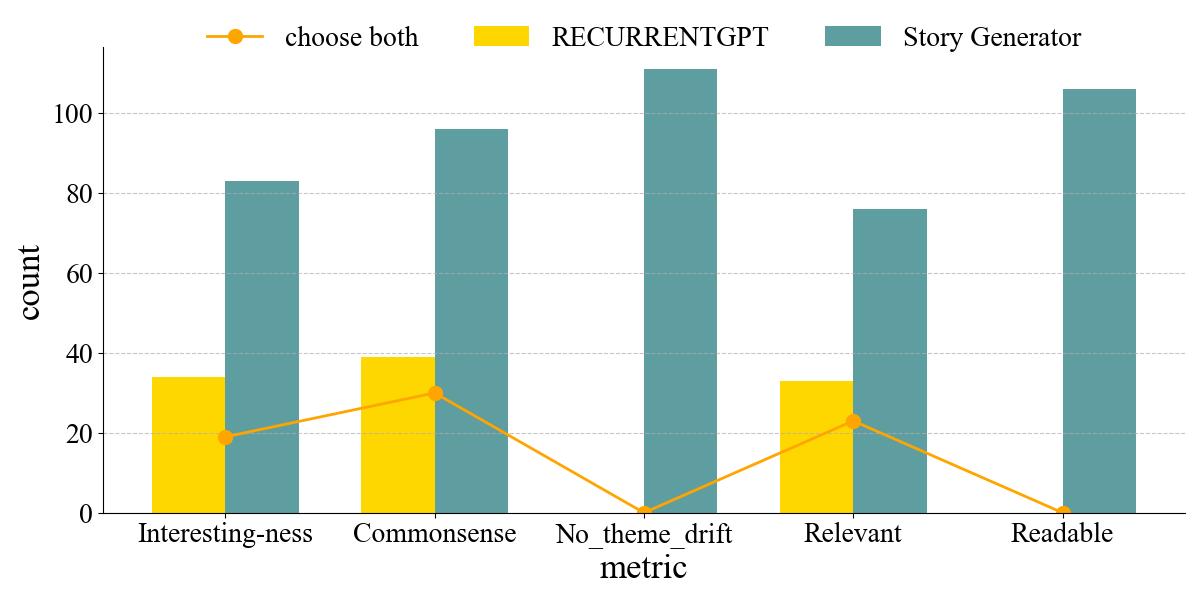}
    \caption{Comparing RECURRENTGPT and Story Generator. Counting identifies certain dimensions of the story.}
    
\centering
    \includegraphics[width=0.3\linewidth]{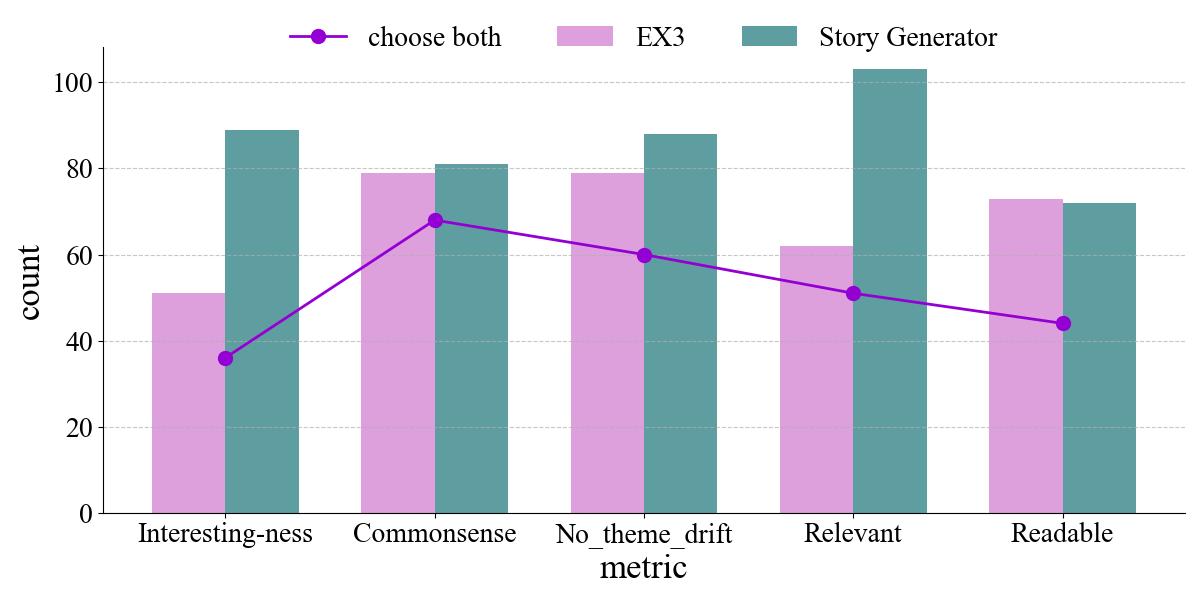}
    \caption{Comparing ${EX^3}$ and Story Generator. Counting identifies certain dimensions of the story.}
\end{figure}

\subsection{Ablation Studies}
The ablation experiments designed for the Story Generator in this paper include two parts: a test on the necessity of the KG-driven obstacle twist plot generator and a test on the necessity of the dialogue between writer simulator and reader simulator.

\begin{figure}[ht]
\centering
\includegraphics[width=0.3\linewidth]{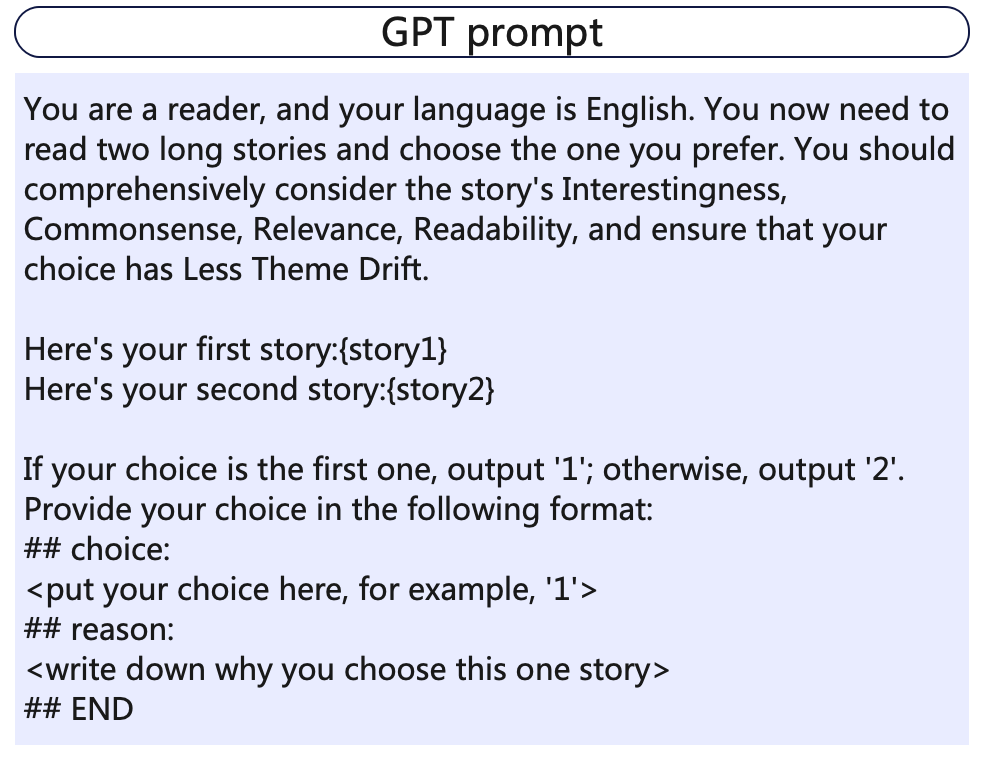} 
\caption{Prompts for GPT to weigh the quality of the stories and select a better one.}
\label{figure7}
\end{figure}

Due to the limited capability of Auto-J-6B and 13B in processing long texts, we use GPT-3.5-turbo to simulate a real human being to select the better story under a set of topics: science fiction, romance, and fantasy. For each topic, 30 stories with each at least 5k words are generated, and the selection of the better stories in the comparison is counted. See task prompt for GPT-3.5-turbo in Figure~\ref{figure7}.

We set up the experimental groups as follows:
\begin{itemize}
    \item \textbf{group 1}: The original Story Generator
    \item \textbf{group 2}: A method that removes the KG-driven twist plot generator. The story outline is entirely generated by the plain story generator
    \item \textbf{group 3}: A method that removes the multi-agent interaction structure of the expander and uses a single agent to expand the text
\end{itemize}

\begin{table}[ht]
\centering

\label{tablename} 
\begin{tabular}{lccc} 
\hline 
 &sci-fi &romance &fantasy  \\ 
\hline
group 1 &18 &15 &24\\
group 2& 12&15 &6\\
\hline
group 1&21&19&22\\
group 3&9&11&8\\
\hline
\end{tabular}
\caption{Better choice counting, total 30 stories' pairs} 
\label{table2}
\end{table}

Through the feedback from the reasons, we have once again confirmed the importance of the twist plot generator and the multi-agent interaction of the expander:

 Story Generator with the twist plot generator can write more interesting plots and introduce more unpredictable characters. In terms of the text scenarios, the emergence of various unexpected events also makes the stories more vivid and full of suspense. Stories with outlines written only by the plain story generator have a certain degree of theme drift, insufficient text interesting-ness, and their detailed descriptions are not as good as those of the original story generator.

The multi-agent expander polishes the sentences, greatly improving the readability of the stories. The emotional logic of the characters is smooth, and the character growth paths are complete. The plots are reasonable and not abrupt, the connections between various plots are strong, and the overall structure and style of the text are relatively consistent.

\section{Analysis and Discussion}

\subsection{Effects of the KG Extraction Method}

For the twist plot generator, extracting KG structure and generating nodes are crucial steps. We need to determine the optimal KG related extraction method. We have designed two methods: one is to use a single agent to extract important KGs from the text, directly generate nodes, and produce new text; the other is to use one agent to extract important information, one agent to abstract KGs, the final one agent to receive KG information, generate nodes and stories. We use other parts of the Story Generator to generate outlines and input them into different twist plot generator configurations. Auto-J is employed as the evaluator. We used pair-wise comparison method to choose the best-performing setting. We generate 30 outlines under each topic for Auto-J to judge which one is better.

\begin{table}[!htbp]
\centering
\label{tablename} 
\begin{tabular}{lccc} 
\hline 
 &sci-fi &romance &fantasy  \\ 
\hline
1 agent &16 &10 &13\\
3 agents&14 &20 &17\\
\hline
\end{tabular}
\label{table3}
\caption{Better choice counting, total 30 stories' pairs} 
\end{table}
We chose 3 agents method to generate twist plots.

\subsection{Effects of Writer and Reader Simulator Dialogue}

Writer and reader simulators are the main agents of the expander part. We want to achieve good text generation results with as few conversation rounds as possible. This round of tests still use Auto-J as evaluator, and Claude-3-7-sonnet as the core LLMs. Defining one writer-reader dialogue as one round, and final Writer editing as the conclusion of the process, we tested 1, 2, 3 rounds to see the performance of iterative optimization. We used Auto-J's single story judging method to choose the best-performing setting. We generate 20 700-word short stories under each topic for Auto-J to rate each one. 

\begin{table}[!htbp]
\centering

\label{tablename} 
\begin{tabular}{lccc} 
\hline 
 &sci-fi &romance &fantasy  \\ 
\hline
1 round& 6.25& 6.7&4.6\\
2 rounds&6.35 & 6.25&4.65\\
3 rounds&4.2 & 5.55&3.15\\
\hline
\end{tabular}
\caption{The average score out of 10, 20 stories under each story topic} 
\label{table4}
\end{table}
We chose 1 round method to generate twist plots.

\subsection{Discussion}
The experimental results demonstrate that the proposed Story Generator effectively addresses the key challenges of long story generation: theme drift and dull plots. By integrating memory mechanisms, literary theory, and multi-agent interaction, our approach outperforms existing baselines across metrics such as interesting-ness, thematic consistency, and readability.  

Firstly, the dual memory storage (long-term and short-term) plays a critical role in mitigating theme drift. Long-term memory anchors the story to its core elements (e.g., topic, main characters, and goals) by continuously summarizing key information, while short-term memory ensures coherence between consecutive segments. This addresses the "Lost in the Middle" phenomenon  by prioritizing critical context retention, unlike outline-based methods that often lose track of initial premises over time. 

Secondly, the KG-driven twist plot framework, inspired by narratology, enhances story appeal by introducing logical yet unexpected obstacles. By constructing knowledge graphs centered on character goals and generating new obstacle nodes, the framework balance creativity and coherence. This contrasts with some general methods that focus on structural expansion but lack explicit mechanisms for thematic twists, resulting in more predictable narratives.

Thirdly, multi-agent interaction between writer and reader simulators improves logical consistency and readability. The dialogue-based revision process mimics human collaborative writing, refining plot holes and ensuring character motivations remain plausible. This addresses limitations of single-agent generation, where long texts often suffer from fragmented logic.  

Together, these elements address key challenges in long story generation. These findings validate the design choices of the Story Generator and underscore the importance of combining memory mechanisms, literary theory, and multi-agent collaboration in advancing long text generation.

\section{Conclusion}

This paper presents Story Generator, a multi-agent structure for long story generation that integrates memory storage and knowledge graphs to address theme drift and generate dull plots, enabling the complete completion of the structural closure. 

Experimental results validate that Story Generator outperforms baselines in human evaluations, with ablation studies confirming the necessity of its core components.  

\bibliographystyle{plain}

\begin{thebibliography}{99}

\bibitem{bae-kim-2024-collective}
Minwook Bae and Hyounghun Kim. 2024.
Collective Critics for Creative Story Generation.
In \textit{Proceedings of the 2024 Conference on Empirical Methods in Natural Language Processing}, pages 18784--18819, Miami, Florida, USA. Association for Computational Linguistics.

\bibitem{Beltagy2020LongformerTL}
Iz Beltagy, Matthew E. Peters, and Arman Cohan. 2020.
Longformer: The Long-Document Transformer.
\textit{ArXiv}, abs/2004.05150.

\bibitem{choi-etal-2025-agents}
Seunghee Choi, Donghoon Kim, and Jaewoo Lee. 2025.
Agents' Room: Narrative Generation through Multi-step Collaboration.
In \textit{Advances in Neural Information Processing Systems 39 (NeurIPS 2025)}, pages 4567--4578. Curran Associates, Inc.

\bibitem{dao2022flashattention}
Tri Dao, Dan Fu, Stefano Ermon, Atri Rudra, and Christopher R{\'e}. 2022.
Flashattention: Fast and memory-efficient exact attention with io-awareness.
\textit{Advances in neural information processing systems}, 35:16344--16359.

\bibitem{doshi2024generative}
Aarti R Doshi and Oliver P Hauser. 2024.
Generative AI enhances individual creativity but reduces the collective diversity of novel content.
\textit{Science Advances}, 10(3):eadn5290.

\bibitem{hochreiter1997long}
Sepp Hochreiter and J{"u}rgen Schmidhuber. 1997.
Long short-term memory.
\textit{Neural Computation}, 9(8):1735--1780.

\bibitem{jiang-etal-2024-leveraging}
Hang Jiang, Xiajie Zhang, Robert Mahari, Daniel Kessler, Eric Ma, Tal August, Irene Li, Alex Pentland, Yoon Kim, Deb Roy, and Jad Kabbara. 2024.
Leveraging Large Language Models for Learning Complex Legal Concepts through Storytelling.
In \textit{Proceedings of the 62nd Annual Meeting of the Association for Computational Linguistics (Volume 1: Long Papers)}, pages 7194--7219, Bangkok, Thailand. Association for Computational Linguistics.

\bibitem{lei-etal-2024-ex3}
Huang Lei, Jiaming Guo, Guanhua He, Xishan Zhang, Rui Zhang, Shaohui Peng, Shaoli Liu, and Tianshi Chen. 2024.
Ex3: Automatic Novel Writing by Extracting, Excelsior and Expanding.
In \textit{Proceedings of the 62nd Annual Meeting of the Association for Computational Linguistics (Volume 1: Long Papers)}, pages 9125--9146, Bangkok, Thailand. Association for Computational Linguistics.

\bibitem{lee-etal-2025-navigating}
Yukyung Lee, Soonwon Ka, Bokyung Son, Pilsung Kang, and Jaewook Kang. 2025.
Navigating the Path of Writing: Outline-guided Text Generation with Large Language Models.
In \textit{Proceedings of the 2025 Conference of the Nations of the Americas Chapter of the Association for Computational Linguistics: Human Language Technologies (Volume 3: Industry Track)}, pages 233--250, Albuquerque, New Mexico. Association for Computational Linguistics.

\bibitem{li2024generative}
Junlong Li, Shichao Sun, Weizhe Yuan, Run-Ze Fan, hai zhao, and Pengfei Liu. 2024.
Generative Judge for Evaluating Alignment.
In \textit{The Twelfth International Conference on Learning Representations}.

\bibitem{liu-etal-2024-lost}
Nelson F. Liu, Kevin Lin, John Hewitt, Ashwin Paranjape, Michele Bevilacqua, Fabio Petroni, and Percy Liang. 2024.
Lost in the Middle: How Language Models Use Long Contexts.
\textit{Transactions of the Association for Computational Linguistics}, 12:157--173.

\bibitem{rashkin-etal-2020-plotmachines}
Hannah Rashkin, Asli Celikyilmaz, Yejin Choi, and Jianfeng Gao. 2020.
PlotMachines: Outline-Conditioned Generation with Dynamic Plot State Tracking.
In \textit{Proceedings of the 2020 Conference on Empirical Methods in Natural Language Processing (EMNLP)}, pages 4274--4295, Online. Association for Computational Linguistics.

\bibitem{tian-etal-2024-large-language}
Yufei Tian, Tenghao Huang, Miri Liu, Derek Jiang, Alexander Spangher, Muhao Chen, Jonathan May, and Nanyun Peng. 2024.
Are Large Language Models Capable of Generating Human-Level Narratives?
In \textit{Proceedings of the 2024 Conference on Empirical Methods in Natural Language Processing}, pages 17659--17681, Miami, Florida, USA. Association for Computational Linguistics.

\bibitem{wang-etal-2025-generating}
Qianyue Wang, Jinwu Hu, Zhengping Li, Yufeng Wang, Daiyuan Li, Yu Hu, and Mingkui Tan. 2025.
Generating Long-form Story Using Dynamic Hierarchical Outlining with Memory-Enhancement.
In \textit{Proceedings of the 2025 Conference of the Nations of the Americas Chapter of the Association for Computational Linguistics: Human Language Technologies (Volume 1: Long Papers)}, pages 1352--1391, Albuquerque, New Mexico. Association for Computational Linguistics.

\bibitem{xie-riedl-2024-creating}
Kaige Xie and Mark Riedl. 2024.
Creating Suspenseful Stories: Iterative Planning with Large Language Models.
In \textit{Proceedings of the 18th Conference of the European Chapter of the Association for Computational Linguistics (Volume 1: Long Papers)}, pages 2391--2407, St. Julian{'}s, Malta. Association for Computational Linguistics.

\bibitem{yang-etal-2022-re3}
Kevin Yang, Yuandong Tian, Nanyun Peng, and Dan Klein. 2022.
Re3: Generating Longer Stories With Recursive Reprompting and Revision.
In \textit{Proceedings of the 2022 Conference on Empirical Methods in Natural Language Processing}, pages 4393--4479, Abu Dhabi, United Arab Emirates. Association for Computational Linguistics.

\bibitem{yang-etal-2023-doc}
Kevin Yang, Dan Klein, Nanyun Peng, and Yuandong Tian. 2023.
DOC: Improving Long Story Coherence With Detailed Outline Control.
In \textit{Proceedings of the 61st Annual Meeting of the Association for Computational Linguistics (Volume 1: Long Papers)}, pages 3378--3465, Toronto, Canada. Association for Computational Linguistics.

\bibitem{zhang2024chain}
Yusen Zhang, Ruoxi Sun, Yanfei Chen, Tomas Pfister, Rui Zhang, and Sercan Arik. 2024.
Chain of agents: Large language models collaborating on long-context tasks.
\textit{Advances in Neural Information Processing Systems}, 37:132208--132237.

\bibitem{10877467}
Nivedita Gupta, M. Vinoth Kumar, Atif Sundake, G. Sharmely, N. Jothi, and A. S. Sivananthavalli. 2024.
AI-Driven Digital Narratives: Revolutionizing Storytelling in Contemporary English Literature through Virtual Reality.
In \textit{2024 Second International Conference Computational and Characterization Techniques in Engineering \& Sciences (IC3TES)}, pages 1--5.

\bibitem{yang2024makesgoodstorymeasure}
Dingyi Yang and Qin Jin. 2024.
What Makes a Good Story and How Can We Measure It? A Comprehensive Survey of Story Evaluation.
\textit{arXiv preprint arXiv:2408.14622}.

\bibitem{gao2025llm}
Mingqi Gao, Xinyu Hu, Xunjian Yin, Jie Ruan, Xiao Pu, and Xiaojun Wan. 2025.
Llm-based nlg evaluation: Current status and challenges.
\textit{Computational Linguistics}, pages 1--27.

\bibitem{sutskever2014sequence}
Ilya Sutskever, Oriol Vinyals, and Quoc V Le. 2014.
Sequence to sequence learning with neural networks.
\textit{Advances in neural information processing systems}, 27.

\bibitem{liu2024memlongmemoryaugmentedretrievallong}
Weijie Liu, Zecheng Tang, Juntao Li, Kehai Chen, and Min Zhang. 2024.
MemLong: Memory-Augmented Retrieval for Long Text Modeling.
\textit{arXiv preprint arXiv:2408.16967}.

\bibitem{forster1927aspects}
E. M. Forster. 1927.
\textit{Aspects of the Novel}. Edward Arnold, London.

\bibitem{zhou2023recurrentgptinteractivegenerationarbitrarily}
Wangchunshu Zhou, Yuchen Eleanor Jiang, Peng Cui, Tiannan Wang, Zhenxin Xiao, Yifan Hou, Ryan Cotterell, and Mrinmaya Sachan. 2023.
RecurrentGPT: Interactive Generation of (Arbitrarily) Long Text.
\textit{arXiv preprint arXiv:2305.13304}.

\bibitem{Todorov1971The2P}
Tzvetan Todorov. 1971.
The 2 Principles of Narrative.
\textit{Diacritics}, 1:37.

\bibitem{jameson1981political}
Fredric Jameson. 1981.
\textit{The Political Unconscious: Narrative as a Socially Symbolic Act}. Cornell University Press.

\bibitem{propp1968morphology}
Vladimir Propp. 1968.
\textit{Morphology of the Folktale}. University of Texas Press. Translated by Laurence Scott.

\end{thebibliography}

\appendix

\section{Appendix: Survey Questionnaire}

Hi! This is a survey that assesses the quality of story writing. You will read settings for two stories, as well as the two corresponding stories we want to compare, A and B. Please read the settings and the content of both stories carefully, and based on your genuine impressions, answer the provided questions below. Thanks for attending.

\subsubsection*{Settings of the two stories}
\begin{itemize}
\item \textbf{Topic}: (example) Love-fiction in high school

\item \textbf{Main character}: (example) Allen and Mika

\item \textbf{Main Goal}: (example) Allen and Mika want to get together forever.
\end{itemize}

\subsection*{Story A}

(example)

Allen first spotted Mika in the hallway between third and fourth period, her backpack slung over one shoulder as she juggled a stack of library books and a half-eaten granola bar. He’d been leaning against his locker, pretending to scroll through his phone, when she stumbled — books teetering, granola bar crumbs scattering like confetti. He moved before he thought, catching Pride and Prejudice before it hit the ground, his fingers brushing hers as she mumbled a flustered “thanks.”

“Jane Austen fan?” he asked, nodding at the book. His voice came out steadier than he felt.

Mika’s cheeks pinked. “My mom says it’s mandatory for anyone who likes stories. I’m starting to see why.”...( $>$10k words)

\subsection*{Story B}

(example)

Allen had always been the quiet one, content to blend into the background of Maryland High's bustling hallways. He found solace in his books and the intricate drawings that filled his sketchpad, rarely looking up to engage with the world around him. But there was one person who always managed to catch his eye: Mika Tanaka.

Mika was the epitome of high school popularity. As the captain of the cheerleading squad and student council president, she seemed to effortlessly navigate the complex social hierarchy of Oakwood High. Her infectious laughter and warm smile drew people to her like moths to a flame, ...( $>$10k words)

\subsection*{Questions}

\begin{enumerate}
    \item After reading these two stories, which one do you think is better?
    \begin{enumerate}
        \item Story A
        \item Story B
    \end{enumerate}

    \item Which story do you think is more appealing and interesting, A or B?
    \begin{enumerate}
        \item Story A
        \item Story B
        \item Both are appealing and interesting.
        \item Neither are appealing and interesting.
    \end{enumerate}

    \item Which story do you think is more logical and rational, A or B?
    \begin{enumerate}
        \item Story A
        \item Story B
        \item Both are logical and rational.
        \item Neither are logical and rational.
    \end{enumerate}

    \item Which story do you think its beginning and ending share a same theme, A or B?
    \begin{enumerate}
        \item Story A
        \item Story B
        \item Both have same theme in their beginning and ending.
        \item Neither have same theme in their beginning and ending.
    \end{enumerate}

    \item Which story do you think is more consistent with the premise, A or B?
    \begin{enumerate}
        \item Story A
        \item Story B
        \item Both are consistent with the premise.
        \item Neither are consistent with the premise.
    \end{enumerate}

    \item Which story do you think is more readable, A or B?
    \begin{enumerate}
        \item Story A
        \item Story B
        \item Both are readable.
        \item Neither are readable.
    \end{enumerate}

\end{enumerate}

\end{document}